\documentclass[letterpaper, 10 pt, conference]{ieeeconf}  

\IEEEoverridecommandlockouts                              

\usepackage{url}      
\usepackage{graphicx}
\usepackage{amsmath}
\usepackage{amssymb}
\usepackage{color,soul}
\usepackage{definitions}
\usepackage{multirow}
\usepackage{siunitx}
\usepackage[absolute,overlay]{textpos}

\usepackage[colorinlistoftodos, german, disable]{todonotes} 

\usepackage[utf8]{inputenc}

\usepackage[font=footnotesize]{caption}
\usepackage[font=footnotesize]{subcaption}

\usepackage{hyperref}

\title{\LARGE \bf
  Adaptive Behavior Generation for Autonomous Driving using Deep Reinforcement Learning with Compact Semantic States
}

\author{Peter Wolf$^{1}$, Karl Kurzer$^{2}$, Tobias Wingert$^{2}$, Florian Kuhnt$^{1}$ and J. Marius Z\"ollner$^{1,2}$ 
  \thanks{
    $^{1}$FZI Research Center for Information Technology, Haid-und-Neu-Str. 10-14, 76131 Karlsruhe, Germany
    {\tt\small \{wolf, kuhnt, zoellner\}@fzi.de}
    $^{2}$Karlsruhe Institute of Technology, Kaiserstr. 12, 76131 Karlsruhe, Germany
    {\tt\small kurzer@kit.edu, wingerttobias@gmail.com}
  }
}

\graphicspath{{03_graphics/}}

\begin{document}
\begin{textblock*}{\textwidth}(19mm,10mm)
	\footnotesize
	\noindent\textcopyright2018~IEEE. Personal use of this material is permitted. Permission from IEEE must be obtained for all other uses, in any current or future media, including reprinting/republishing this material for advertising or promotional purposes, creating new collective works, for resale or redistribution to servers or lists, or reuse of any copyrighted component of this work in other works.\\
	\textit{2018 IEEE Intelligent Vehicles Symposium (IV), pp. 993-1000, 26-30 June 2018}
\end{textblock*}
\maketitle
\thispagestyle{empty}
\pagestyle{empty}

\begin{abstract}
Making the right decision in traffic is a challenging task that is highly dependent on individual preferences as
well as the surrounding environment.
Therefore it is hard to model solely based on expert knowledge.
In this work we use Deep Reinforcement Learning to learn maneuver decisions based on a compact semantic state representation.
This ensures a consistent model of the environment across scenarios as well as a behavior adaptation function, enabling on-line changes
of desired behaviors without re-training.
The input for the neural network is a simulated object list similar to that of Radar or Lidar
sensors, superimposed by a relational semantic scene
description.
The state as well as the reward are extended by a behavior adaptation function and a parameterization respectively.
With little expert knowledge and a set of mid-level actions, it can be seen that the agent is capable
to adhere to traffic rules and learns to drive safely in a variety of situations.

\end{abstract}

\section{Introduction}
\label{s:introduction}

While sensors are improving at a staggering pace and actuators as well as control theory are well up to par to the
challenging task of autonomous driving, it is yet to be seen how a robot can devise decisions that navigate it safely in
a heterogeneous environment that is partially made up by humans who not always take rational decisions or have known cost
functions.

Early approaches for maneuver decisions focused on predefined rules embedded in large state machines, each requiring
thoughtful engineering and expert knowledge \cite{ROB:ROB20255, ROB:ROB20248, ROB:ROB20262}.

Recent work focuses on more complex models with additional domain knowledge to predict and generate maneuver decisions \cite{Ulbrich2015}.
Some approaches explicitly model the interdependency between the actions of traffic participants \cite{Lawitzky2013}
as well as address their replanning capabilities \cite{Bahram2016}.

With the large variety of challenges that vehicles with a higher degree of autonomy need to face, the limitations of
rule- and model-based approaches devised by human expert knowledge that proved successful in the past become apparent.

At least since the introduction of AlphaZero, which discovered the same game-playing strategies as humans did in Chess
and Go in the past, but also learned entirely unknown strategies, it is clear, that human expert knowledge is
overvalued \cite{Silver2017,Silver2017a}.
Hence, it is only reasonable to apply the same techniques to the task of behavior planning in autonomous driving,
relying on data-driven instead of model-based approaches.

The contributions of this work are twofold.
First, we employ a compact semantic state representation, which is based on the most significant relations between other
entities and the ego vehicle.
This representation is neither dependent on the road geometry nor the number of surrounding vehicles, suitable for a
variety of traffic scenarios.
Second, using parameterization and a behavior adaptation function we demonstrate the ability to train agents with a
changeable desired behavior, adaptable on-line, not requiring new training.

\begin{figure}
  \centering
    \def\svgwidth{0.8\columnwidth}
    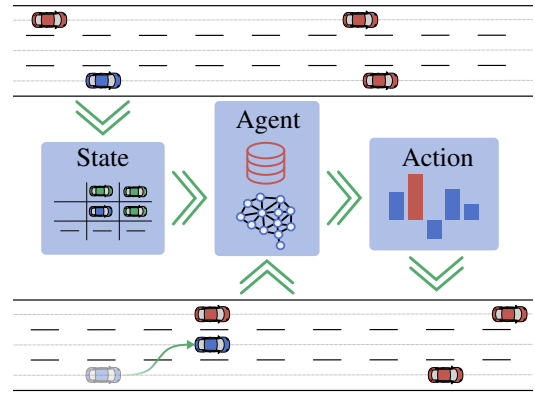
    \caption{The initial traffic scene is transformed into a compact semantic state representation $s$ and used as input
    for the reinforcement learning (RL) agent.
    The agent estimates the action $a$ with the highest return (Q-value) and executes it, e.g., changing lanes.
    Afterwards a reward $r$ is collected and a new state $s'$ is reached.
    The transition $(s, a, r, s')$ is stored in the agent's replay memory.}
  \label{fig:overview}
\end{figure}

The remainder of this work is structured as follows:
In Section~\ref{s:related_work} we give a brief overview of the research on behavior generation in the automated driving
domain and deep reinforcement learning.
A detailed description of our approach, methods and framework follows in Section~\ref{s:approach} 
and Section~\ref{s:experiments} respectively.
In Section~\ref{s:evaluation} we present the evaluation of our trained agents.
Finally, we discuss our results in Section~\ref{s:conclusion}.

\section{Related Work}
\label{s:related_work}
Initially most behavior planners were handcrafted state machines, made up by a variety of modules to handle different
tasks of driving.
During the DARPA Urban Challenge Boss (CMU) for example used five different modules to conduct on road driving.
The responsibilities of the modules ranged from lane selection, merge planning to distance keeping \cite{ROB:ROB20255}.
Other participants such as Odin (Virginia Tech) or Talos (MIT) developed very similar behavior generators
\cite{ROB:ROB20248, ROB:ROB20262}.

Due to the limitations of state machines, current research has expanded on the initial efforts
by creating more complex and formal models:
A mixture of POMDP, stochastic non-linear MPC and domain knowledge can be used to generate lane 
change decisions in a variety of traffic scenarios \cite{Ulbrich2015}.
Capturing the mutual dependency of maneuver decisions between different agents, planning can be 
conducted with foresight \cite{Lawitzky2013, Bahram2016}.
While \cite{Lawitzky2013} plans only the next maneuver focusing on the reduction of collision probabilities between all
traffic participants, \cite{Bahram2016} explicitly addresses longer planning horizons and the replanning capabilities
of others.

Fairly different is the high-level maneuver planning presented by \cite{Kohlhaas2015} that is based on \cite{Kohlhaas2014a},
where a semantic method to model the state and action space of the ego vehicle is introduced.
Here the ego vehicle's state is described by its relations to relevant objects of a structured traffic scene, such as
other vehicles, the road layout and road infrastructure.
Transitions between semantic states can be restricted by traffic rules, physical barriers or other constraints.
Combined with an A* search and a path-time-velocity planner a sequence of semantic states and their implicit
transitions can be generated.

In recent years, deep reinforcement learning (DRL) has been successfully used to learn policies for various challenges.
Silver et al. used DRL in conjunction with supervised learning on human game data to train the policy networks of their
program AlphaGo~\cite{Silver2016}; \cite{Kober2013, Li2017} present an overview of RL and DRL respectively.
In \cite{Silver2017} and \cite{Silver2017a} their agents achieve superhuman performance in their respective domains
solely using a self-play reinforcement learning algorithm which utilizes Monte Carlo Tree Search (MCTS) to accelerate the
training.
Mnih et al. proposed their deep Q-network (DQN)~\cite{mnih2013playing,Mnih2015} which was able to learn policies for a plethora of
different Atari 2600 games and reach or surpass human level of performance.
The DQN approach offers a high generalizability and versatility in tasks with high dimensional state spaces and has been
extended in various work~\cite{Schaul2015,VanHasselt2015,Wang2016}. 
Especially actor-critic approaches have shown huge success in learning complex policies and are also able to learn behavior
policies in domains with a continuous action space~\cite{Mnih2016,wu2017scalable}.

In the domain of autonomous driving, DRL has been used to directly control the movements of simulated vehicles to solve
tasks like lane-keeping~\cite{sallab2017deep,wolf2017learning}.
In these approaches, the agents receive sensor input from the respective simulation environments and are trained to
determine a suitable steering angle to keep the vehicle within its driving lane.
Thereby, the focus lies mostly on low-level control.

In contrast, DRL may also be applied to the problem of forming long term driving strategies for different driving
scenarios.
For example, DRL can be used in intersection scenarios to determine whether to cross an
intersection by controlling the
longitudinal movement of the vehicle along a predefined path~\cite{Isele2017}.
The resulting policies achieve a lower wait time than using a Time-To-Collision policy.
In the same way, DRL techniques can be used to learn lateral movement, e.g. lane change maneuvers in a highway
simulation~\cite{Mirchevska2017}.

Since it can be problematic to model multi-agent scenarios as a Markov Decision Process (MDP) due to the unpredictable
behavior of other agents, one possibility is to decompose the problem into learning a cost function for driving
trajectories~\cite{Shalev-Shwartz2016a}.
To make learning faster and more data efficient, expert knowledge can be incorporated by 
restricting certain actions during the training process~\cite{mukadam2017tactical}.
Additionally, there is the option to handle task and motion planning by learning low-level controls for lateral as well
as longitudinal maneuvers from a predefined set and a high-level maneuver policy~\cite{Paxton2017a}.

Our approach addresses the problem of generating adaptive behavior that is applicable to a 
multitude of traffic scenarios using a semantic state representation.
Not wanting to limit our concept to a single scenario, we chose a more general set of ``mid-level'' 
actions, e.g. \emph{decelerate} instead of \emph{Stop} or \emph{Follow}.
This does not restrict the possible behavior of the agent and enables the agent to identify 
suitable high-level actions implicitly.
Further, our agent is trained without restrictive expert knowledge, but rather we let 
the agent learn from its own experience, allowing it to generate superior solutions without the 
limiting biases introduced by expert knowledge \cite{Silver2017,Silver2017a}.

\section{Approach}
\label{s:approach}

\begin{figure}
	\vspace{0.7em}
	\centering
	\begin{subfigure}{\columnwidth}
		\includegraphics[width=\columnwidth]{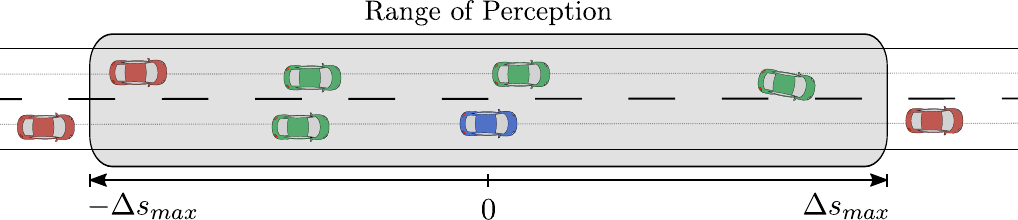}
		\caption{The ego vehicle (blue) is driving on a two lane road. Five other vehicles are in 
			its sensor range. Based on
			a vehicle scope with $\Lambda_{lateral} = \Lambda_{ahead} = \Lambda_{behind} = 1$ only 
			four vehicles (green) are
			considered in the semantic state representation.}
		\label{fig:range_of_perception}
	\end{subfigure}\hfill%
	\begin{subfigure}{\columnwidth}
		\centering
		\def\svgwidth{0.7\columnwidth}
		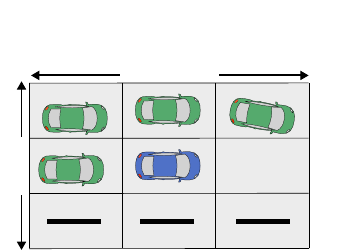
		\caption{The relational grid resulting from the scene of (a) with
			$\Lambda_{lateral} = \Lambda_{ahead} = \Lambda_{behind} = 1 $.
			Moreover, there is no adjacent lane to the right, which is represented in the grid as 
			well.}
		\label{fig:relational_grid}
	\end{subfigure}%
	\caption{An example scene (a) is transformed to a relational grid (b)
		using a vehicle scope $\Lambda$ with $\Lambda_{lateral} = \Lambda_{ahead} = 
		\Lambda_{behind} = 1 $.
		The red vehicle, which is in sensor range, is not represented in the grid.
		Since there is no vehicle driving on the same lane in front of the ego vehicle,
		the respective relational position in the grid is empty.
	}
	\label{fig:relational_grid_full}
\end{figure}

We employ a deep reinforcement learning approach to generate adaptive behavior for autonomous driving.

A reinforcement learning process is commonly modeled as an MDP~\cite{sutton1998reinforcement} $(S, A, R, \delta, T)$
where $S$ is the set of states, $A$ the set of actions, $R: S\times A\times S \rightarrow \mathbb{R}$ the reward
function, $\delta: S\times A \rightarrow S$ the state transition model and $T$ the set of terminal states.
At timestep $i$ an agent in state $s \in S$ can choose an action $a \in A$ according to a policy $\pi$ and will progress
into the successor state $s'$ receiving reward $r$.
This is defined as a transition $t = (s, a, s', r)$.

The aim of reinforcement learning is to maximize the future discounted return $G_i = \sum_{n=i} 
\gamma^{n-i} r_i$.
A DQN uses Q-Learning~\cite{watkins1992q} to learn Q-values for each action given input state $s$ based on past transitions.
The predicted Q-values of the DQN are used to adapt the policy $\pi$ and therefore change the agent's behavior.
A schematic of this process is depicted in Fig.~\ref{fig:overview}.

For the input state representation we adapt the ontology-based concept from \cite{Kohlhaas2014a} focusing on relations with other
traffic participants as well as the road topology.
We design the state representation to use high level preprocessed sensory inputs such as object
lists provided by
common Radar and Lidar sensors and lane boundaries from visual sensors or map data.
To generate the desired behavior the reward is comprised of different factors with varying priorities.
In the following the different aspects are described in more detail.

\subsection{Semantic Entity-Relationship Model}
\begin{figure}
	\vspace{0.7em}
	\centering
	\includegraphics[width=\columnwidth]{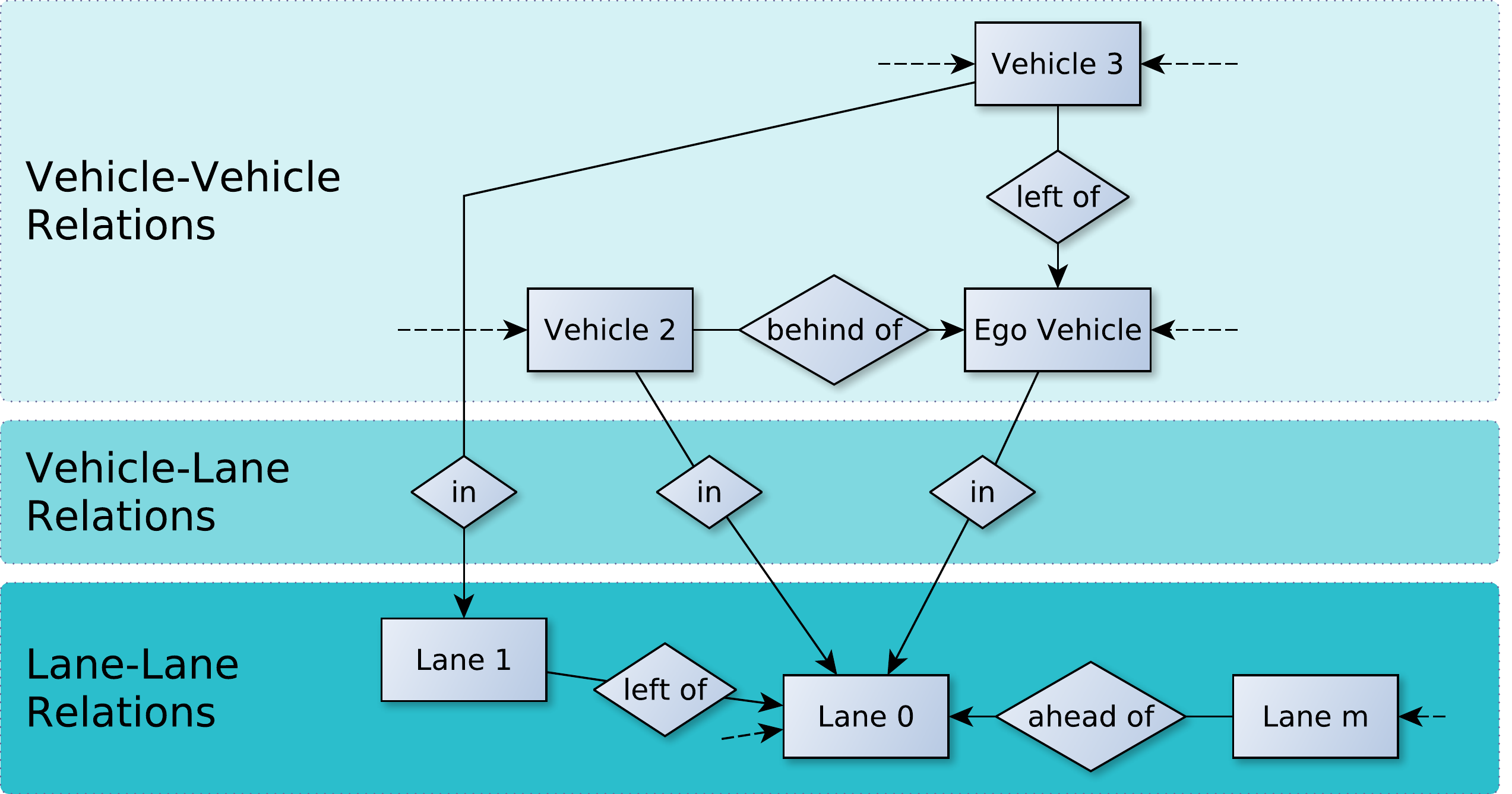}
	\caption{Partially visualized entity-relationship model of the scene in 
	Fig.~\ref{fig:range_of_perception}.
		The vehicle topology is modeled by vehicle-vehicle relations whereas the lane topology is 
		modeled by lane-lane relations.
	}
	\label{fig:entity_relationships}
\end{figure}

A traffic scene $\trafficScene$ is described by a semantic entity-relationship model, consisting of all scene objects
and relations between them.
We define it as the tuple $(\mathcal{E}, \mathcal{R})$, where
\begin{itemize}
  \setlength\itemsep{0.1em}
  \item $\mathcal{E} = \{e_0, e_1, ..., e_n \}$: set of scene objects (entities).
  \item $\mathcal{R} = \{r_0, r_1, ..., r_m \}$: set of relations.
\end{itemize}

The scene objects contain all static and dynamic objects, such as vehicles, pedestrians, lane segments, signs and traffic lights.

In this work we focus on \emph{vehicles} $\mathcal{V} \subset \mathcal{E}$, \emph{lane segments}
$\mathcal{L} \subset \mathcal{E}$ and the three relation types  \emph{vehicle-vehicle relations}, \emph{vehicle-lane relations}
and \emph{lane-lane relations}.
Using these entities and relations an entity-relationship representation of a traffic scene can be created as depicted in
Fig.~\ref{fig:entity_relationships}.
Every and relation holds several properties or attributes of the scene objects, such as e.g. 
absolute positions or
relative velocities.

This scene description combines low level attributes with high level relational knowledge in a generic way.
It is thus applicable to any traffic scene and vehicle sensor setup, making it a
beneficial state representation.

But the representation is of varying size and includes more aspects than are relevant for a given driving task.
In order to use this representation as the input to a neural network we transform it
to a fixed-size relational grid that includes only the most relevant relations.

\subsection{Relational Grid}

We define a relational grid, centered at the ego vehicle $\ego \in \mathcal{V}$, see Fig. \ref{fig:relational_grid_full}.
The rows correspond to the relational lane topology, whereas the columns correspond to the vehicle topology on these lanes.

To define the size of the relational grid, a vehicle scope $\Lambda$ is introduced that captures the lateral and
longitudinal dimensions, defined by the following parameters:

\begin{itemize}
  \setlength\itemsep{0.1cm}
  \item $\Lambda_{lateral} \in \mathbb{N}$: maximum number of laterally adjacent lanes to the ego vehicle's lane
    that is considered
  \item $\Lambda_{ahead} \in \mathbb{N}$: maximum number of vehicles per lane driving in front of the ego vehicle
    that is considered
  \item $\Lambda_{behind} \in \mathbb{N}$: maximum number of vehicles per lane driving behind the ego vehicle that
    is considered
\end{itemize}

The set of vehicles inside the vehicle scope $\Lambda$ is denoted by $\mathcal{V}_{\Lambda} \subseteq \mathcal{V}$.
The different features of vehicles and lanes are represented by separate layers of the grid, 
resulting in a semantic state representation, see Fig.~\ref{fig:state_space}.
The column vector of each grid cell holds attributes of the vehicle and the lane it belongs to.
While the ego vehicle's features are absolute values, other vehicles' features are relative to the ego vehicle or the
lane they are in (induced by the vehicle-vehicle and vehicle-lane relations of the 
entity-relationship model):

\subsubsection{Non-ego vehicle features}

The features of non-ego vehicles
$v_{i} \in \mathcal{V}_{\Lambda} \backslash \ego$ are $\boldsymbol{f}_{i}^\vehicles = (\Delta s_i, \Delta \dot{s}_i, \Delta d_i, \Delta \phi_i)$:

\begin{itemize}
	\setlength\itemsep{0.1cm}
	\item $\Delta s_i$: longitudinal position relative to \ego\
	\item $\Delta \dot{s}_i$: longitudinal velocity relative to \ego\
	\item $\Delta d_i$: lateral alignment relative to lane center
	\item $\Delta\phi_i$: heading relative to lane orientation
\end{itemize}

\subsubsection{Ego vehicle features}

To generate adaptive behavior, the ego vehicle's features include a function $\Omega(\tau, \theta)$ that
describes the divergence from the desired behavior of the ego vehicle parameterized by $\theta$ given the traffic scene
$\tau$.

Thus the features for $v_{ego}$ are $\boldsymbol{f}_{ego}^v = (\Omega(\tau, \theta), \dot{s}_{ego}, k)$:
\begin{itemize}
	\setlength\itemsep{0.1cm}
	\item $\Omega(\tau,\theta)$: behavior adaptation function
	\item $\dot{s}_{ego}$: longitudinal velocity of ego vehicle
	\item $k$: lane index of ego vehicle\\($k = 0$ being the right most lane)
\end{itemize}

\subsubsection{Lane features}

The lane topology features of $l_{k} \in \lanes$ are $\boldsymbol{f}_{k}^\lanes = (\laneType, \Delta s_{k})$:
\begin{itemize}
	\setlength\itemsep{0.1cm}
	\item $\laneType$: lane type (normal, acceleration)
	\item $\Delta s_{k}$: lane ending relative to \ego
\end{itemize}

Non-perceivable features or missing entities are indicated in the semantic state representation by 
a dedicated value.
The relational grid ensures a consistent representation of the environment, independent of the road geometry or
the number of surrounding vehicles.

\begin{figure}
  \vspace{0.9em}
  \centering
  \def\svgwidth{0.8\columnwidth}
  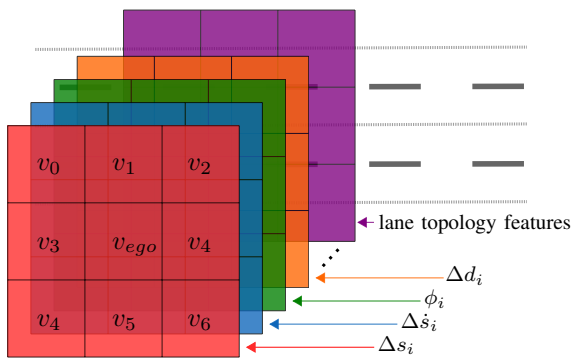
  \caption{The relational grid contains one layer per feature.
    The vehicle features $\boldsymbol{f}_i^\vehicles$ and $\boldsymbol{f}_{ego}^\vehicles$ share layers and  are in the
    cells of the $v_i$ and \ego\ respectively.
    The lane features $\boldsymbol{f}_k^\lanes$ are on additional layers in the $k$-th row of the 
    grid.
  }
  \label{fig:state_space}
\end{figure}

The resulting input state $s \in S$ is depicted in Fig.~\ref{fig:state_space} and fed into a DQN.

\subsection{Action Space}
The vehicle's actions space is defined by a set of semantic actions that is deemed sufficient for 
most on-road driving
tasks, excluding special cases such as U-turns.
The longitudinal movement of the vehicle is controlled by the actions \textit{accelerate} and \textit{decelerate}.
While executing these actions, the ego vehicle keeps its lane.
Lateral movement is generated by the actions \textit{lane change left} as well as \textit{lane change right} respectively.
Only a single action is executed at a time and actions are executed in their entirety, the vehicle is not able to prematurely
abort an action.
The \textit{default} action results in no change of either velocity, lateral alignment or heading.

\subsection{Adaptive Desired Behavior through Reward Function}

With the aim to generate adaptive behavior we extend the reward function $R(s,a)$ by a parameterization $\theta$.
This parameterization is used in the behavior adaptation function $\Omega(\tau,\theta)$, so that the agent is able to
learn different desired behaviors without the need to train a new model for varying parameter values.

Furthermore, the desired driving behavior consists of several individual goals, modeled by separated rewards.
We rank these reward functions by three different priorities.
The highest priority has collision avoidance, important but to a lesser extent are rewards associated with traffic rules,
least prioritized are rewards connected to the driving style.

The overall reward function $R(s,a, \theta)$ can be expressed as follows:

\begin{equation}
  \label{eq:reward_general}
  R(s,a,\theta) = \begin{cases}
  R_{collision}(s, \theta) 		& s \in S_{collision} \\
  R_{rules}(s, \theta)			& s \in S_{rules}\\
  R_{driving\ style}(s,a, \theta)	& \text{else},
  \end{cases}
\end{equation}

The subset $S_{collision} \subset S$ consists of all states $s$ describing a collision state of the
ego vehicle $v_{ego}$ and another vehicle $v_{i}$.
In these states the agent only receives the immediate reward without any possibility to earn any other future rewards.
Additionally, attempting a lane change to a nonexistent adjacent lane is also treated as a collision.

The state dependent evaluation of the reward factors facilitates the learning process.
As the reward for a state is independent of rewards with lower priority, the eligibility trace is more concise for the
agent being trained.
For example, driving at the desired velocity does not mitigate the reward for collisions.

\section{Experiments}
\label{s:experiments}

\subsection{Framework}
\begin{figure}[h]
  \centering
  \includegraphics[width=\columnwidth]{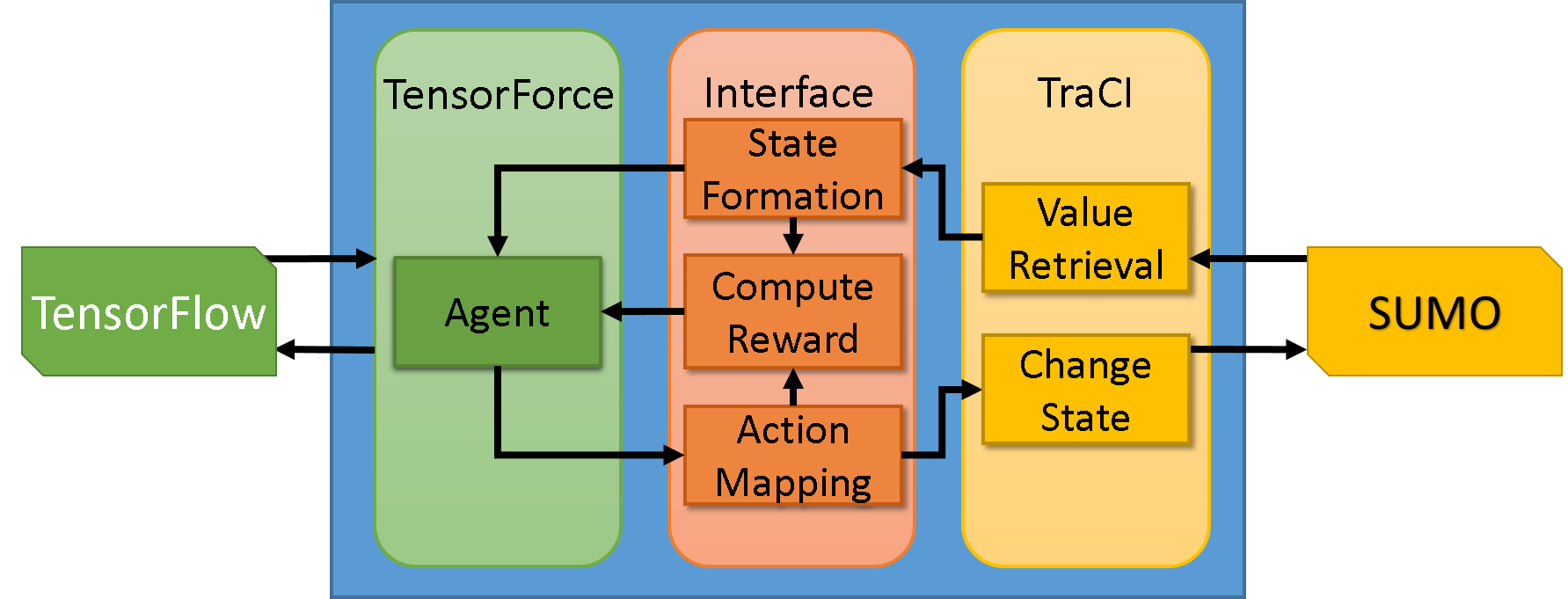}
  \caption{
    The current state from SUMO is retrieved and transformed into the semantic state representation.
    This is the input to the RL agent, which is trained using the TensorForce library.
    Chosen actions are mapped to a respective state change of the simulation.
    The agent's reward is then computed based on the initial state, the chosen action and the 
    successor state.
  }
  \label{fig:framework}
\end{figure}

While our concept is able to handle data from many preprocessing methods used in autonomous vehicles, we tested the
approach with the traffic simulation SUMO~\cite{Krajzewicz2012}.
A schematic overview of the framework is depicted in Fig. \ref{fig:framework}.
We use SUMO in our setup as it allows the initialization and execution of various traffic scenarios with adjustable
road layout, traffic density and the vehicles's driving behavior.
To achieve this, we extend TensorForce~\cite{schaarschmidt2017tensorforce} with a highly configurable interface
to the SUMO environment.
TensorForce is a reinforcement library based on TensorFlow~\cite{tensorflow2015-whitepaper}, which enables the
deployment of various customizable DRL methods, including DQN.

\begin{figure*}
  \vspace{0.7em}
  \centering
  \begin{subfigure}{\columnwidth}
  \includegraphics[width=\columnwidth]{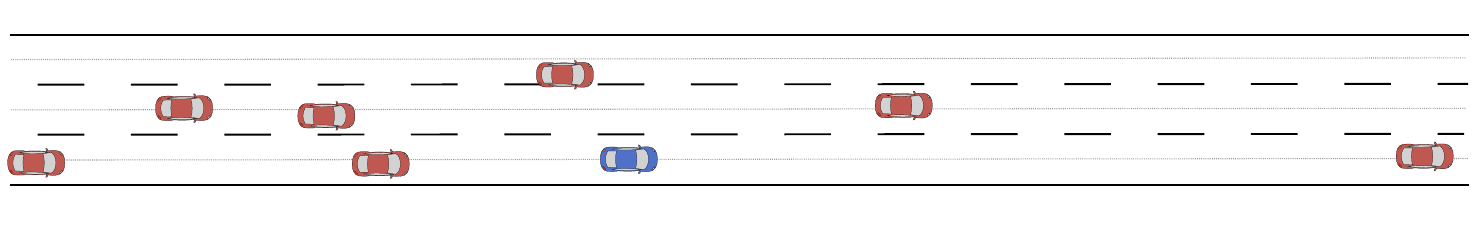}%
  \caption{Highway Driving}%
  \label{fig:highway_driving}%
  \end{subfigure}\hfill%
  \begin{subfigure}{\columnwidth}
  \includegraphics[width=\columnwidth]{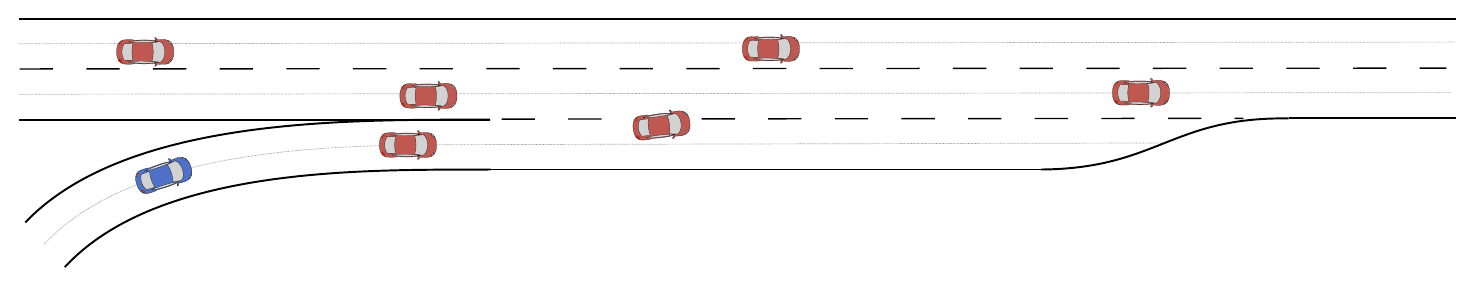}%
  \caption{Highway Merging}%
  \label{fig:highway_merging}%
  \end{subfigure}%
  \caption{To examine the agent's compliance to traffic rules, it is trained and evaluated on two different traffic
    scenarios.
    In (a) the agent has the obligation to drive on the right most lane and must not pass others
    from the right, amongst other constraints.
    In (b) the agent is allowed to accelerate while on the on-ramp and also might overtake
    vehicles on its left. But it has to leave the on-ramp before it ends.
  }
  \label{fig:scenarios}
\end{figure*}

The interface fulfills three functions: the state extraction from SUMO, the calculation of the reward and the mapping of
the chosen maneuver onto valid commands for the SUMO controlled ego vehicle.
The necessary information about the ego vehicle, the surrounding vehicles as well as additional information about the
lanes are extracted from the running simulation.
These observed features are transformed into the state representation of the current scene.
For this purpose the surrounding positions relative to the ego vehicle are checked and if another vehicle fulfills a relation,
the selected feature set for this vehicle is written into the state representation for its relation.
The different reward components can be added or removed, given a defined value or range of values and a priority.

This setup allows us to deploy agents using various DRL methods, state representations and rewards in multiple traffic
scenarios.
In our experiments we used a vehicle scope with $\Lambda_{behind} = 1$ and $\Lambda_{ahead} = \Lambda_{lateral} = 2$.
This allows the agent to always perceive all lanes of a 3 lane highway and increases its potential anticipation.

\subsection{Network}

In this work we use the DQN approach introduced by Mnih et al.~\cite{Mnih2015} as it has shown its capabilities
to successfully learn behavior policies for a range of different tasks.
While we use the general learning algorithm described in~\cite{Mnih2015}, including the usage of experience replay and a
secondary target network, our actual network architecture differs from theirs.
The network from Mnih et al. was designed for a visual state representation of the environment.
In that case, a series of convolutional layers is commonly used to learn a suitable low-dimensional feature set from
this kind of high-dimensional sensor input.
This set of features is usually further processed in fully-connected network layers.

Since the state representation in this work already consists of selected features, the learning of a low-dimensional
feature set using convolutional layers is not necessary. 
Therefore we use a network with solely fully-connected layers, see Tab.~\ref{tab:network_architecture}. 

The size of the input layer depends on the number of features in the state representation.
On the output layer there is a neuron for each action.
The given value for each action is its estimated Q-value.

\begin{table}[h]
  \centering
  \begin{tabular}{|c|c|}
    \hline
    Layer & Neurons \\
    \hline
    Input  & $-$ \\
    Hidden Layer 1& $512$ \\
    Hidden Layer 2& $512$ \\
    Hidden Layer 3& $256$ \\
    Hidden Layer 4& $64$ \\
    Output& $5$ \\
    \hline
  \end{tabular}
  \caption{The network layout of our DQN agent. The size of the input
           neurons is determined by the number of features in the state space. The
           five output neurons predict the five $Q$-values for respective actions.}
  \label{tab:network_architecture}
\end{table}

\subsection{Training}

During training the agents are driving on one or more traffic scenarios in SUMO.
An agent is trained for a maximum of 2 million timesteps, each generating a transition consisting of the observed state,
the selected action, the subsequent state and the received reward.
The transitions are stored in the replay memory, which holds up to 500,000 transitions.
After reaching a threshold of at least 50,000 transitions in memory, a batch of 32 transitions is randomly selected to
update the network's weights every fourth timestep.
We discount future rewards by a factor of $0.9$ during the weight update.
The target network is updated every 50,000th step.

To allow for exploration early on in the training, an $\epsilon$-greedy policy is followed.
With a probability of $\epsilon$ the action to be executed is selected randomly, otherwise the action with the highest
estimated Q-value is chosen.
The variable $\epsilon$ is initialized as $1$, but decreased linearly over the course of 500,000 timesteps until it
reaches a minimum of $0.1$, reducing the exploration in favor of exploitation.
As optimization method for our DQN we use the RMSProp algorithm~\cite{hinton2012rmsprop} with a learning rate of $10^{-5}$
and decay of $0.95$.

The training process is segmented into episodes.
If an agent is trained on multiple scenarios, the scenarios alternate after the end of an episode.
To ensure the agent experiences a wide range of different scenes, it is started with a randomized departure time, lane,
velocity and $\theta$ in the selected scenario at the beginning and after every reset of the simulation.
In a similar vein, it is important that the agent is able to observe a broad spectrum of situations in the scenario
early in the training.
Therefore, should the agent reach a collision state $s \in S_{collision}$ by either colliding with another
vehicle or attempting to drive off the road, the current episode is finished with a terminal state.
Afterwards, a new episode is started immediately without reseting the simulation or changing the agent's position or velocity.
Since we want to avoid learning an implicit goal state at the end of a scenario's course, the simulation is reset if a
maximum amount of 200 timesteps per episode has passed or the course's end has been reached and the episode ends with a
non-terminal state.

\subsection{Scenarios}

\begin{figure*}[!ht]
	\centering
	\begin{minipage}[b]{.98\columnwidth}
		\includegraphics[width=0.98\columnwidth]{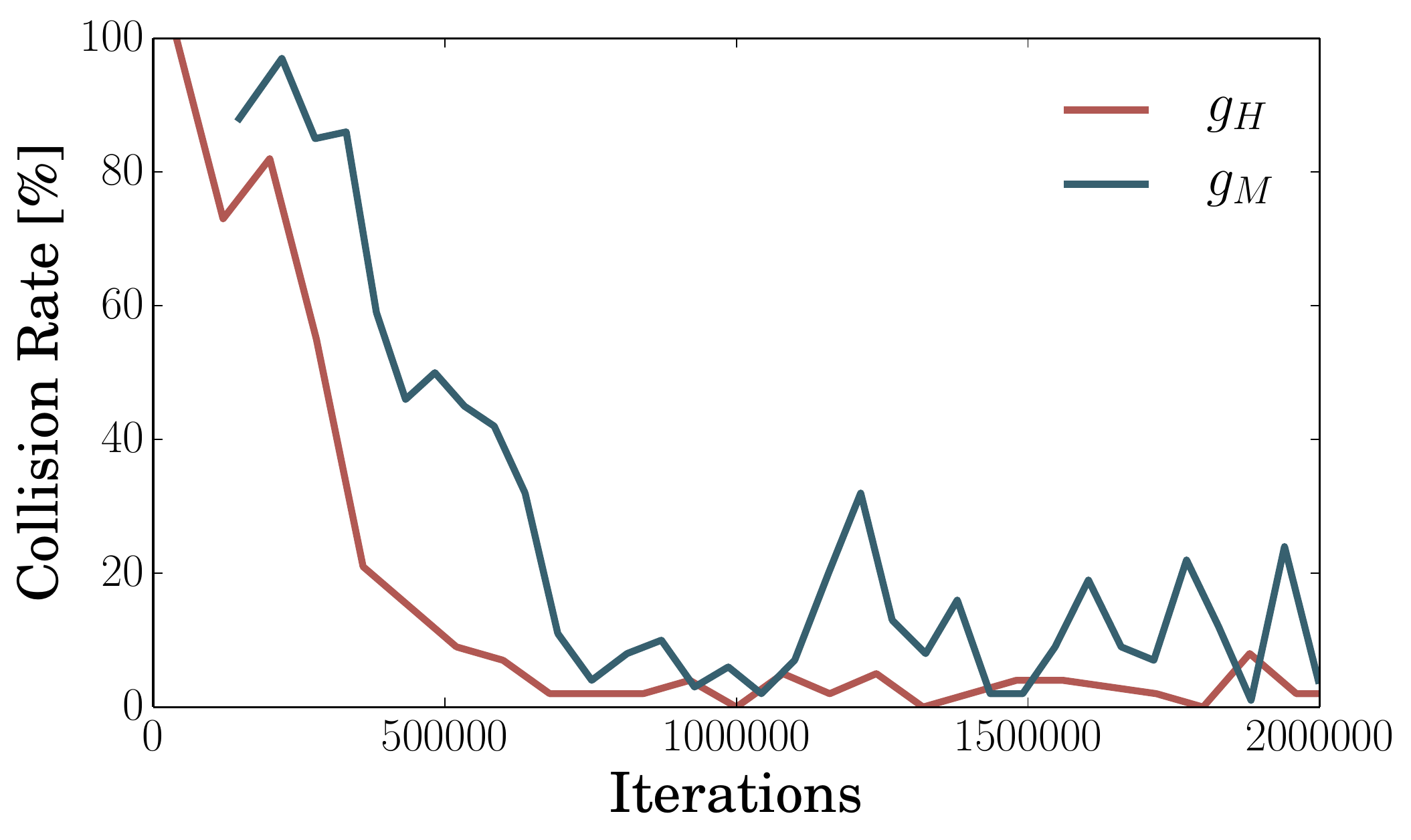}
		\caption{Collision Rate of the agents $g_H$ and $g_M$ during training.}
		\label{fig:collision_rate}
		\vspace{1em}
	\end{minipage}\hfill
	\begin{minipage}[b]{.98\columnwidth}
		\vspace{0em}
		\includegraphics[width=0.98\columnwidth]{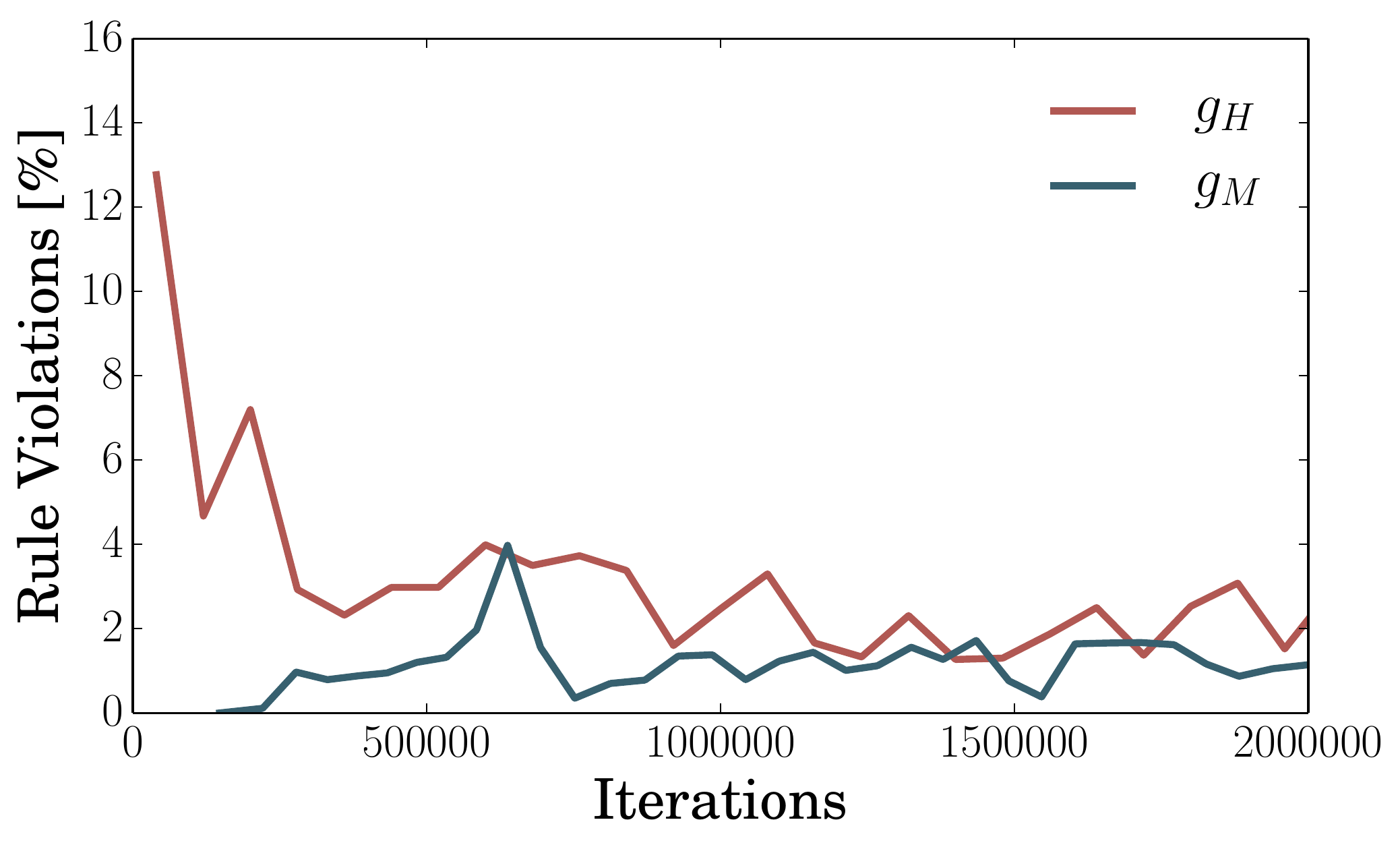}
		\caption{Relative duration of rule violations of the agents $g_H$ and $g_M$ during 
			training.}
		\label{fig:rule_violations}
	\end{minipage}
\end{figure*}

Experiments are conducted using two different scenarios, see Fig. \ref{fig:scenarios}.
One is a straight multi-lane highway scenario.
The other is a merging scenario on a highway with an on-ramp.

To generate the desired adaptive behavior, parameterized reward functions are defined (see Eq.~\ref{eq:reward_general}).
We base $R_{rules}$ on German traffic rules such as the obligation to drive on the right most lane ($r_{keepRight}$),
prohibiting overtaking on the right ($r_{passRight}$) as well as keeping a minimum distance to vehicles in front
($r_{safeDistance}$).
A special case is the acceleration lane, where the agent is allowed to pass on the right and is not required to stay on
the right most lane.
Instead the agent is not allowed to enter the acceleration lane ($r_{notEnter}$).

Similarly, $R_{driving\ style}$ entails driving a desired velocity ($r_{velocity}$) ranging from \SI{80}{km/h} to
\SI{115}{km/h} on the highway and from \SI{40}{km/h} to \SI{80}{km/h} on the merging scenario.
The desired velocity in each training episode is defined by $\theta_v$ which is sampled uniformly over the
scenario's velocity range.
Additionally, $R_{driving\ style}$ aims to avoid unnecssary lane and velocity changes ($r_{action}$).

With these constraints in mind, the parameterized reward functions are implemented as follows, to produce the desired
behavior.

\begin{equation}
R_{collision}(s, \theta) = \theta_{t} r_{collision}
\label{eq:r_terminal}
\end{equation}

\begin{equation}
  \begin{split}
    R_{rules}(s, \theta) &= r_{passRight}(s,\theta_{p}) + r_{notEnter}(s,\theta_{n})\\
    &+ r_{safeDistance}(s,\theta_{s}) + r_{keepRight}(s,\theta_{k} )
    \end{split}
  \label{eq:r_rules}
\end{equation}

\begin{equation}
  \begin{split}
    R_{driving\ style}(s, a, \theta) &= r_{action}(a,\theta_a) + r_{velocity}(s,\theta_v)
    \end{split}
  \label{eq:r_driving_style}
\end{equation}

To enable different velocity preferences, the behavior adaptation function $\Omega$ returns the 
difference between the desired and the actual velocity of \ego.

\section{Evaluation}
\label{s:evaluation}

During evaluation we trained an agent $g_H$ only on the highway scenario and an agent $g_M$ only on
the merging scenario.
In order to show the versatility of our approach, we additionally trained an agent $g_C$ both on the
highway as well as the merging scenario (see Tab.\ref{tab:results}).
Due to the nature of our compact semantic state representation we are able to achieve this without
further modifications.
The agents are evaluated during and after training by running the respective scenarios 100 times.
To assess the capabilities of the trained agents, using the concept mentioned in Section
\ref{s:approach}, we introduce the following metrics.

\textbf{Collision Rate [\%]:} The collision rate denotes the average amount of collisions over all test runs.
In contrast to the training, a run is terminated if the agent collides.
As this is such a critical measure it acts as the most expressive gauge assessing the agents performance.

\textbf{Avg. Distance between Collisions [km]:} The average distance travelled between collisions is used to remove the
bias of the episode length and the vehicle's speed.

\textbf{Rule Violations [\%]:} Relative duration during which the agent is not keeping a safe distance or is overtaking
on right.

\textbf{Lane Distribution [\%]:} The lane distribution is an additional weak indicator for the agent's compliance with
the traffic rules.

\textbf{Avg. Speed [m/s]:} The average speed of the agent does not only indicate how fast the agent drives, but also
displays how accurate the agent matches its desired velocity.

\begin{figure}[h]
	\centering
	\includegraphics[width=1.0\columnwidth]{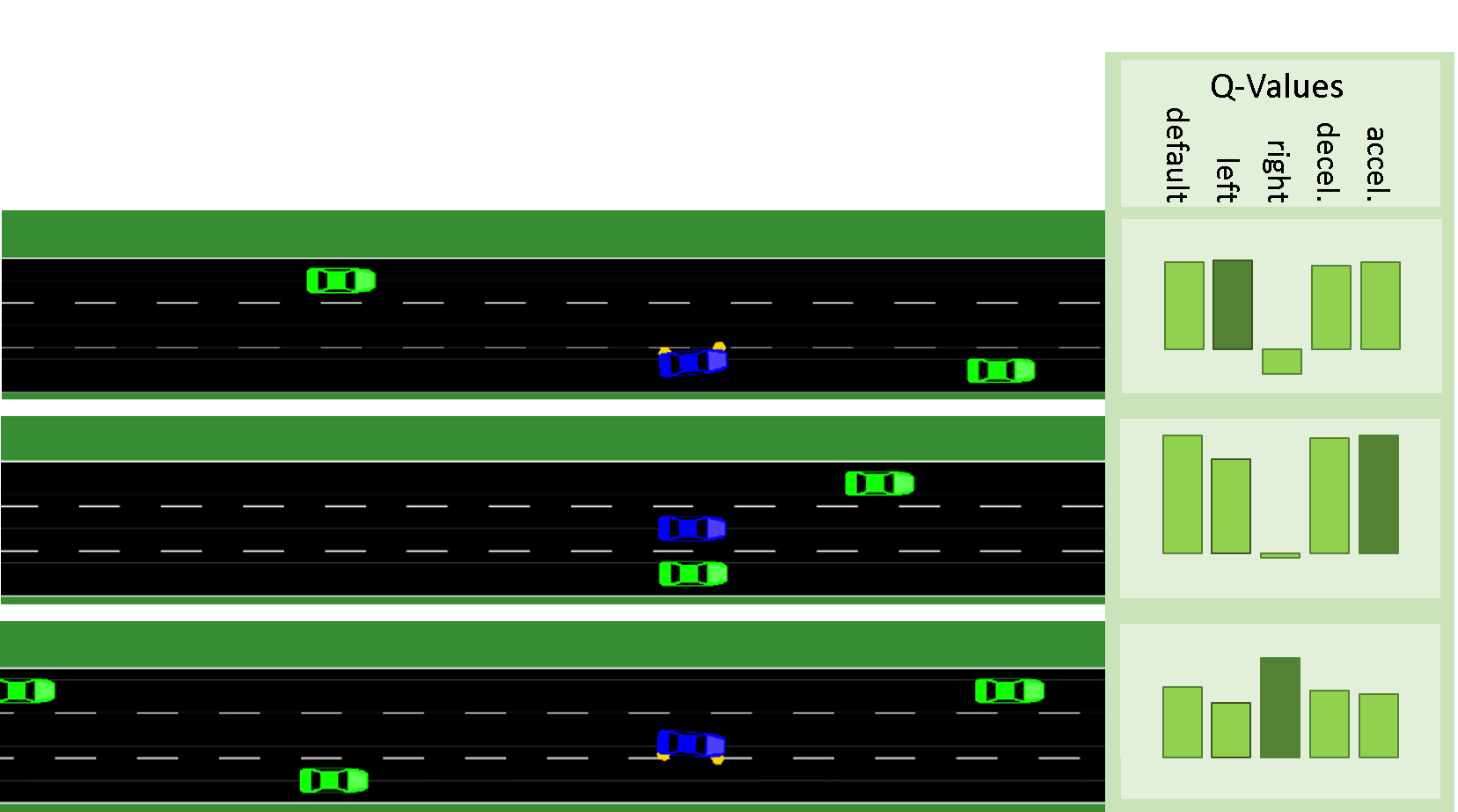}
	\caption{The agent (blue) is driving behind a slower vehicle (green).
		In this situation the action \emph{lane change left} has the highest estimated Q-value.
		After the lange change, the agent accelerates and overtakes the slower vehicle.
		Subsequently, the agent changes back to the right most lane.}
	\label{fig:lanechange}
	\vspace{1em}
\end{figure}

The results of the agents trained on the different scenarios are shown in Tab.~\ref{tab:results}.
The agents generally achieve the desired behavior.
An example of an overtaking maneuver is presented in Fig.~\ref{fig:lanechange}.
During training the collision rate of $g_H$ decreases to a decent level (see 
Fig.~\ref{fig:collision_rate}).
Agent $g_M$ takes more training iterations to reduce its collision rate to a reasonable level, as it not only has to
avoid other vehicles, but also needs to leave the on-ramp.
Additionally, $g_M$ successfully learns to accelerate to its desired velocity on the on-ramp.
But for higher desired velocities this causes difficulties leaving the ramp or braking in time in case the middle lane
is occupied.
This effect increases the collision rate in the latter half of the training process.

The relative duration of rule violations by $g_H$ reduces over the course of the training, but 
stagnates at
approximately $2\%$ (see Fig.~\ref{fig:rule_violations}).
A potential cause is our strict definition of when an overtaking on the right occurs.
The agent almost never performs a full overtaking maneuver from the right, but might drive faster than another
vehicle on the left hand side, which will already be counted towards our metric.
For $g_M$ the duration of rule violations is generally shorter, starting low, peaking and then also stagnating similarly
to $g_H$.
This is explained due to overtaking on the right not being considered on the acceleration lane.
The peak emerges as a result of the agent leaving the lane more often at this point.

The lane distribution of $g_H$ (see Tab.~\ref{tab:results}) demonstrates that the agent most often drives on the right
lane of the highway, to a lesser extent on the middle lane and only seldom on the left lane.
This reflects the desired behavior of adhering to the obligation of driving on the right most lane and only using the
other lanes for overtaking slower vehicles.
In the merging scenario this distribution is less informative since the task does not provide the same opportunities for
lane changes.

\begin{table}[h]
  \vspace{0.6em}
  \centering
  \setlength\tabcolsep{4.0pt}
    \caption{Results of the trained agents. The agents $g_H$ and $g_M$ were only evaluated on their 
    respective scenario,
    	while $g_C$ was evaluated on both.
    	The results for $g_C$ are listed separately for each scenario.
    }
  \begin{tabular}{|l|r|r|r|r|}
    \hline
    & \multicolumn{1}{|c|}{$g_H$} & \multicolumn{1}{|c|}{$g_M$} & \multicolumn{1}{|c|}{$g_C$ (highway)} &
    \multicolumn{1}{|c|}{$g_C$ (merging)} \\
    \hline
    Collision Rate & $2\%$ & $1\%$ & $2\%$ & $4\%$ \\
    \hline
    avg. Distance  & \SI{98.37}{km} & \SI{28.98}{km} & \SI{75.92}{km} & \SI{7.19}{km} \\
    \hline
    Rule Violations & $1.52\%$ & $0.87\%$ & $1.62\%$ & $0.5\%$ \\
    \hline
    Lane 0 & $64.12\%$ & $87.99\%$ & $57.11\%$ & $87.8\%$ \\
    \hline
    Lane 1 & $22.07\%$ & $11.81\%$ & $34.46\%$ & $11.85\%$ \\
    \hline
    Lane 2 & $13.81\%$ & $0.2\%$ & $8.43\%$ & $0.35\%$ \\
    \hline
  \end{tabular}
  \label{tab:results}
\end{table}

To measure the speed deviation of the agents, additional test runs with fixed values for the desired velocity were
performed.
The results are shown in Tab.~\ref{tab:speed_dev}.
As can be seen, the agents adapt their behavior, as an increase in the desired velocity raises the average speed of the
agents.
In tests with other traffic participants, the average speed is expectedly lower than the desired velocity, as the agents
often have to slow down and wait for an opportunity to overtake.
Especially in the merging scenario the agent is unable to reach higher velocities due to these circumstances.
During runs on an empty highway scenario, the difference between the average and desired velocity diminishes.

Although $g_H$ and $g_M$ outperform it on their individual tasks, $g_C$ achieves satisfactory 
behavior on both.
Especially, it is able to learn task specific knowledge such as overtaking in the acceleration lane of the on-ramp while
not overtaking from the right on the highway.

A video of our agents behavior is provided online.\footnote{http://url.fzi.de/behavior-iv2018}

\section{Conclusions}
\label{s:conclusion}

In this work two main contributions have been presented.
First, we introduced a compact semantic state representation that is applicable to a variety of traffic scenarios.
Using a relational grid our representation is independent of road topology, traffic constellation and sensor setup.

Second, we proposed a behavior adaptation function which enables changing desired driving behavior online without the
need to retrain the agent.
This eliminates the requirement to generate new models for different driving style preferences or other varying
parameter values.

\begin{table}[h]
  \vspace{0.6em}
  \centering
  \setlength\tabcolsep{3.0pt}
    \caption{The average speed of each agent given varying desired velocities [\SI{}{m/s}].
    	The agents have been evaluated on the training scenarios with normal traffic density as 
    	well as on an empty highway.
    	While not every agent is able to achieve the desired velocity precisely, their behavior 
    	adapts to the different
    	parameter values.
    }
  \begin{tabular}{|c|c|c|c|c|c|c|c|}
    \hline
    \multirow{1}{*}{} &
    \multicolumn{2}{|c|}{$g_H$} &
    \multicolumn{2}{|c|}{$g_M$} &
    \multicolumn{3}{|c|}{$g_C$} \\
    \hline
    $v_{des}$ & highway & empty & merging & empty & highway & merging & empty \\
    \hline
    12& $12.8$ & $13.2$ & $10.8$ & $13.3$ & $12.1$ & $10.3$ & $12.6$ \\
    \hline
    17& $17.1$ & $17.1$ & $13.5$ & $18.6$ & $16.9$ & $13.2$ & $17.1$ \\
    \hline
    22& $21.1$ & $21.9$ & $14.7$ & $23.7$ & $20.9$ & $14.5$ & $21.8$ \\
    \hline
    25& $23.4$ & $24.8$ & $15.1$ & $26.5$ & $23.0$ & $15.0$ & $23.9$ \\
    \hline
    30& $25.7$ & $29.2$ & $15.3$ & $30.4$ & $25.2$ & $14.8$ & $28.0$ \\
    \hline
  \end{tabular}
  \label{tab:speed_dev}
\end{table}

Agents trained with this approach performed well on different traffic scenarios, i.e. highway driving and highway
merging.
Due to the design of our state representation and behavior adaptation, we were able to develop a single model
applicable to both scenarios.
The agent trained on the combined model was able to successfully learn scenario specific behavior.

One of the major goals for future work we kept in mind while designing the presented concept is the transfer from the
simulation environment to real world driving tasks.
A possible option is to use the trained networks as a heuristic in MCTS methods, similar to~\cite{Paxton2017a}.
Alternatively, our approach can be used in combination with model-driven systems to plan or evaluate driving behavior.

To achieve this transfer to real-world application, we will apply our state representation to further traffic scenarios,
e.g., intersections.
Additionally, we will extend the capabilities of the agents by adopting more advanced reinforcement learning
techniques.

\section*{Acknowledgements}
\label{s:acknowledgements}
We wish to thank the German Research Foundation (DFG) for funding the project Cooperatively Interacting Automobiles
(CoInCar) within which the research leading to this contribution was conducted.
The information as well as views presented in this publication are solely the ones expressed by the
authors.

\listoftodos

\addtolength{\textheight}{-10.8cm}   


\bibliographystyle{IEEEtran}
\bibliography{IEEEabrv,04_mendeley-export/library}

\end{document}